\documentclass[letterpaper]{article} 
\usepackage{aaai25}  
\usepackage{times}  
\usepackage{helvet}  
\usepackage{courier}  
\usepackage[hyphens]{url}  
\usepackage{graphicx} 
\usepackage{natbib}  
\usepackage{caption} 
\usepackage{algorithm}
\usepackage{algorithmic}
\usepackage{latexsym}
\usepackage{arydshln}
\usepackage{booktabs}
\usepackage{multirow}
\usepackage{graphicx}
\usepackage{subfigure}
\usepackage{amsmath}
\usepackage{amssymb}
\usepackage{booktabs}
\usepackage{multirow}
\usepackage{array}
\usepackage{bm}
\usepackage{xparse}
\usepackage{mathtools}
\usepackage{bm}
\usepackage{amssymb}
\usepackage{multirow}
\usepackage{booktabs}
\usepackage{array}
\usepackage{pifont}
\usepackage{enumitem}
\usepackage{subfigure}
\usepackage{colortbl}
\usepackage{colortbl}
\usepackage[table,xcdraw]{xcolor}
\nocopyright

\usepackage{subcaption}
\usepackage{newfloat}
\usepackage{listings}
\DeclareCaptionStyle{ruled}{labelfont=normalfont,labelsep=colon,strut=off} 
\floatstyle{ruled}
\newfloat{listing}{tb}{lst}{}
\floatname{listing}{Listing}
\title{\textbf{CITI}: Enhancing Tool Utilizing Ability in Large Language Models without \\ Sacrificing General Performance}
\author{
    Yupu Hao\textsuperscript{\rm 1, \rm2}, 
    Pengfei Cao\textsuperscript{\rm 1, \rm2}, 
    Zhuoran Jin\textsuperscript{\rm 1, \rm2}, 
    Huanxuan Liao\textsuperscript{\rm 1, \rm2}, 
    Yubo Chen\textsuperscript{\rm 1, \rm2}, \\
    Kang Liu\textsuperscript{\rm 1, \rm2}, 
    Jun Zhao\textsuperscript{\rm 1, \rm2}
}
\affiliations{
    \textsuperscript{\rm 1} The Laboratory of Cognition and Decision Intelligence for Complex Systems, \\Institute of Automation, Chinese Academy of Sciences, Beijing, China\\
    \textsuperscript{\rm 2} School of Artificial Intelligence, University of Chinese Academy of Sciences, Beijing, China\\

    \{haoyupu2023, liaohuanxuan2023\}@ia.ac.cn,
    \{pengfei.cao, zhuoran.jin, yubo.chen, kliu, jzhao\}@nlpr.ia.ac.cn

}

\usepackage{bibentry}

\begin{document}

\maketitle

\begin{abstract}
Tool learning enables the Large Language Models (LLMs) to interact with the external environment by invoking tools, enriching the accuracy and capability scope of LLMs. However, previous works predominantly focus on improving model's tool-utilizing accuracy and the ability to generalize to new, unseen tools, excessively forcing LLMs to adjust specific tool-invoking pattern without considering the harm to model's general performance. This deviates from the actual applications and original intention of integrating tools to enhance model. To tackle this problem, we dissect the capability trade-offs by examining the hidden representation changes and the gradient-based importance score of model's components. Based on the analysis result, we propose a \textbf{C}omponent \textbf{I}mportance-based \textbf{T}ool-utilizing ability \textbf{I}njection method (\textbf{CITI}). According to the gradient-based importance score of different components, it alleviates the capability conflicts caused by fine-tuning process by applying distinct training strategies to different components. \textbf{CITI} applies Mixture-Of-LoRA (MOLoRA) for important components. Meanwhile, it fine-tunes the parameters of few components deemed less important in the backbone of the LLM, while keeping other parameters frozen. \textbf{CITI} can effectively enhance the model's tool-utilizing capability without excessively compromising its general performance. Experimental results demonstrate that our approach achieves outstanding performance across a range of evaluation metrics.\footnote{Code is available at \url{https://github.com/hypasd-art/CITI}}
\end{abstract}

%

\section{Introduction}

Large Language Models (LLMs) have demonstrated significant capabilities in understanding language and generating human-like text \cite{zhao2023survey}. Despite their impressive performance, LLMs still face limitations, such as the tendency to generate hallucinated information \cite{huang2023survey} and an inability to interact with the real physical world \cite{qin2023tool}. To tackle these limitations and expand the model's capabilities scope beyond traditional natural language tasks, there is a growing interest in enabling LLMs to interact with the external environment by equipping various tools \cite{qin2024toolllm, schick2023toolformer, gao2023palprogramaidedlanguagemodels, chen2023program, nakano2022webgpt,lu2023chameleonplugandplaycompositionalreasoning}. 

\begin{figure}[t]
  \includegraphics[width=1\linewidth]{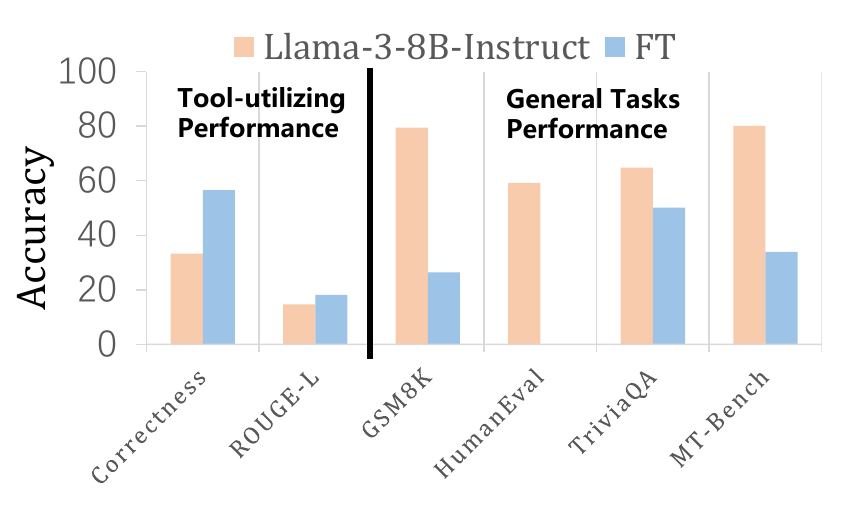} 
  \caption {The model's performance in tool-utilizing and general tasks, FT represents full parameter fine-tuning on dataset API-Bank \cite{li-etal-2023-api}.  
  }
    \label{fig:apibank_performance}
\end{figure}

There have been a substantial number of benchmarks evaluating different tool utilizing aspects \cite{ye2024tooleyes, li-etal-2023-api, zhan2024injecagent}, along with a suite of proposed methodologies \cite{qin2024toolllm,tang2023toolalpaca}. These approaches predominantly leverage in-context learning \cite{shen2023hugginggpt, wu2023visual} or fine-tuning techniques \cite{shen2024small, gao2023confucius,chen2024advancingtoolaugmentedlargelanguage} to equip the model with tool-invoking ability. In contrast to in-context learning, fine-tuning has gained popularity for its capability to deeply integrate task-specific knowledge into model parameters, enhancing its adaptability to specific tasks. However, as illustrated in Figure~\ref{fig:apibank_performance}, we find that excessive fine-tuning on tool learning datasets severely diminishes the model's broader cognitive competencies, leading to a decline in the model's general abilities and catastrophic forgetting about parametric knowledge. 

Tools serve models rather than models serving tools. Consequently, it is crucial for the tool-augmented models to incorporate tool-utilizing ability while simultaneously preserving their original general versatility. Thus, an intriguing question is \textit{what happens in LLMs when fine-tuning on tool learning dataset?} We dissect the question from two aspects: hidden representation and components. In detail, \textbf{from hidden representation perspective}, when subtracting the original hidden representations from those representations tuned on tool learning dataset, we observe a significant phenomenon: the increments, derived by subtraction, exhibit a \textbf{Co-directional shift} in the hidden state space, which implies there is a strong correlation in the direction of increments between various tool-unrelated tasks (e.g. mathematics, coding) and tool-related tasks.  
\textbf{From components perspective}, for different capabilities, we calculate the gradient-based importance score ranking of model's linear modules (referred to as the linear components) based on corresponding datasets respectively. We discover that the important linear components in the ranking play more vital role in corresponding ability expression while the unimportant components contribute less to this ability. Concurrently, we find that the importance rankings of different abilities are highly consistent. This suggests that certain components are pivotal in the expression of the model's abilities across a broad range of tasks, whereas some components exert a more limited influence on most tasks. And we further explore the impact of fine-tuning different components based on tool-related importance ranking. It is found that optimizing important components results in a significant decrease in general performance, while the model may not fully learn the knowledge of calling tools only by optimizing unimportant components.

Based on the insights from our analysis, we propose a novel \textbf{C}omponent \textbf{I}mportance-based \textbf{T}ool-utilizing ability \textbf{I}njection method (\textbf{CITI}), which applies Mixture-of-LoRA adapters to important components and adopts full parameters fine-tuning on unimportant components. Firstly, we identify the gradient-based importance score of all the linear components in the model. Secondly, for the important linear components, we integrate a set of Mixture-Of-LoRA (MOLoRA) adapters to absorb the tool-invoking knowledge. To handle the Co-directional shift phenomenon, we design a router network to separate the tool-related and tool-unrelated inputs, reducing the impact on model's backbone. Thirdly, for the unimportant linear components, we employ full parameter fine-tuning to take full advantage of more parameters. 
Specifically, Our training process adopts a three-stage approach. In the initial training stage, \textit{Router Pre-training}, we pre-train the router network in MOLoRA to teach it to distinguish tool-related and tool-unrelated inputs. In the second stage, \textit{MOLoRA Improvement}, we concentrate on fine-tuning the MOLoRA adapters while freezing the backbone of LLM. In the third stage, \textit{Unimportant Components Optimization}, \textbf{CITI} fine-tunes a small part of unimportant components within the backbone to improve model's performance while maintaining its general abilities.

We conduct abundant experiments on tool learning datasets, including API-Bank \cite{li-etal-2023-api} and ToolAlpaca \cite{tang2023toolalpaca}, and evaluate model's preservation of general abilities involved mathematics, code generation, factual knowledge and instruction following. The results show great effectiveness of our method. 

To summarize, the contributions of our work include:
\begin{itemize}
    \item We discover the phenomenon that fine-tuning LLMs on tool learning datasets significantly influence the general abilities, and we have conducted a thorough inspect of this trade-off phenomenon through analysing the perspective of hidden representation and components. Adequate experiments and analysis demonstrate the factors resulting the catastrophic forgetting of general abilities.
    \item We propose a \textbf{C}omponent \textbf{I}mportance-based \textbf{T}ool-utilizing ability \textbf{I}njection method (\textbf{CITI}), alleviating the trade-offs by applying distinct strategies to different components identified by gradient-based importance score of the model.
    \item Our method achieves competitive tool-utilizing results across two tool learning datasets. Additionally, it preserves general performance which is 7.59\% superior than LoRA and 31.95\% superior than full parameters fine-tuning in average in dataset API-Bank, and it is also 8.96\% better than LoRA and 29.03\% better than full parameters fine-tuning in dataset ToolAlpaca, demonstrating significant effectiveness to enhance tool-utilizing ability while preserving LLMs general performance. 
\end{itemize}

\section{Analysis of Hidden Representation and Components}

Hidden representation and components play a crucial role in determining the hidden state of inputs as they propagate from lower layers to higher layers, helping us understand the inner working process of LLM.
Specifically, we conduct the following analytical experiments and assess general abilities on a range of frequently-used datasets, including GSM8K (GSM) \cite{cobbe2021training}, HumanEval (HE) \cite{chen2021evaluating}, TriviaQA (TQA) \cite{joshi2017triviaqa}, MT-Bench (MT) \cite{NEURIPS2023_91f18a12}. 

\subsection{Hidden Representation Perspective}
\label{sec:analysis hidden representation}
The hidden representation refers to the output vector by the each layer of the auto-regressive transformer model. Our goal is to explore the changes in these representation following fine-tuning with the tool learning instructions. 

\subsubsection{Analytical Approach}
We introduce the concept of Incremental Change of Capability ($ICC$) to quantify the changes in the hidden representation of fine-tuning process.
Specifically, for ability $t$ and instruction $m$ within a processed dataset $\mathcal{DM}_t$, $H_{REF}^l(m)$ is the original hidden representation of the last token in instruction $m$ in untrained model at layer $l$, while $H_{SFT}^l(m)$ represents the representation of the same token after fine-tuning. And the $ICC_t^l$ can be represented as:

\begin{equation}
\footnotesize
ICC_t^l = \frac{1}{|\mathcal{DM}_t|} \sum_{m \in \mathcal{DM}_t} H_{SFT}^l(m) - H_{REF}^l(m)
\end{equation}

The average of vector's increment computed by $\mathcal{DM}_t$ is considered to represent the changes in hidden space of ability $t$. The cosine similarity of these increments is then computed to assess the relationship of vector's change between ability $a$ and $b$ in model's hidden state space:

\begin{equation}
\bm{Sim}(a, b) = \frac{ICC_a \cdot ICC_b}{|ICC_a| \times |ICC_b|} 
\end{equation}

\subsubsection{Experimental Results}

To get representations, we randomly truncate a portion of the golden answer and append it after the instruction, creating a ``final input''. For a particular task, we sample 1000 ``final input'' from the dataset to construct $\mathcal{DM}_t$ and put them into the model. The vector corresponding to last token of the ``final input'' $m$ of different layers is considered as the hidden representation vector $H(m)$. Then we compute the similarity of different $ICC$.

\begin{figure}[htbp]
  \includegraphics[width=\columnwidth]{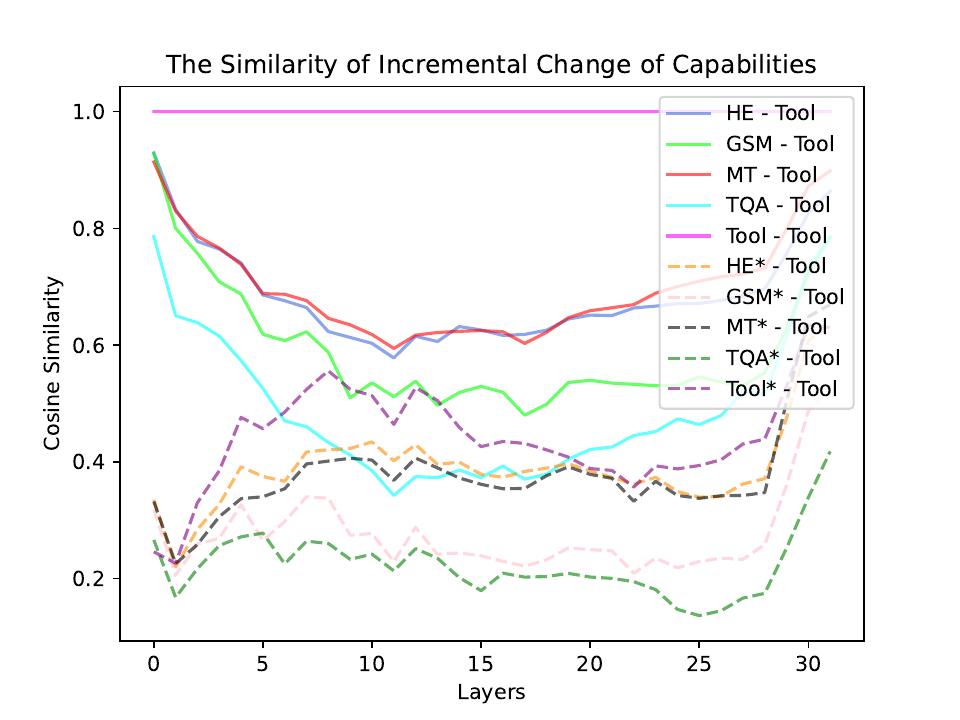}
  \caption{Cosine similarity of $ICC$ between the input of different layers of Feed-Forward Network (FFN) in model Meta-Llama-3-8B-Instruct, where the notation with an asterisk (*) represents $ICC$ fine-tuned on the code-related dataset (e.g. TQA* represents $ICC$ of TriviaQA trained by code dataset), and no asterisk (*) represents $ICC$ fine-tuned on tool learning dataset.}
  \label{fig:sim_mlp}
\end{figure}

In Figure~\ref{fig:sim_mlp}, the solid lines represent the similarity between tool-related increments $ICC_{tool}$ with general task increments $ICC_{t}$, where LLM is fine-tuned on tool learning dataset. For better comparison, we train a model on a code generation dataset and computed its increments $ICC_t *$. Here $t$ represents a specific general ability.

We find that \textbf{there is a \textbf{Co-directional shift} phenomenon in model's hidden state space.} The changing direction of the $ICC_t$ on the general task is positively correlated with the direction of $ICC_{tool}$ in model's hidden state space, comparing with the dashed line. 
Our hypothesis is that after fine-tuning by tool learning datasets, the model cannot distinguish tool-related and tool-unrelated inputs in the hidden state space correctly, resulting in more similar activation states for different types of instructions, which affects the general performance of the model.

\begin{figure*}[htbp]
	\centering
	\begin{minipage}{0.32\linewidth}
		\centering
		\includegraphics[width=1\linewidth]{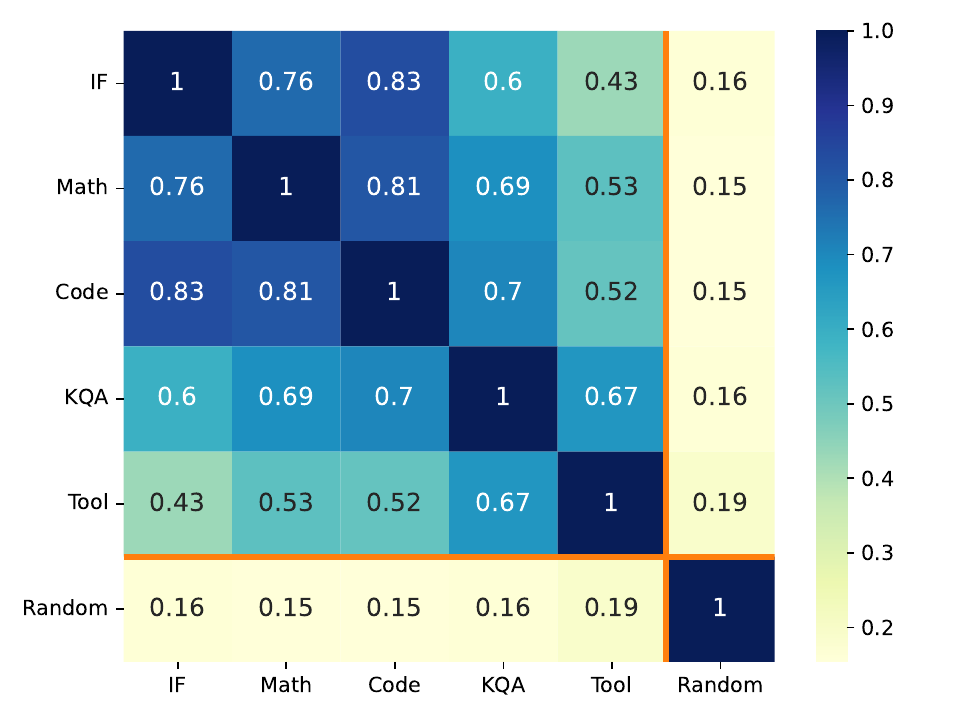}
		\caption*{high importance groups}

	\end{minipage}
    \begin{minipage}{0.32\linewidth}
		\centering
		\includegraphics[width=1\linewidth]{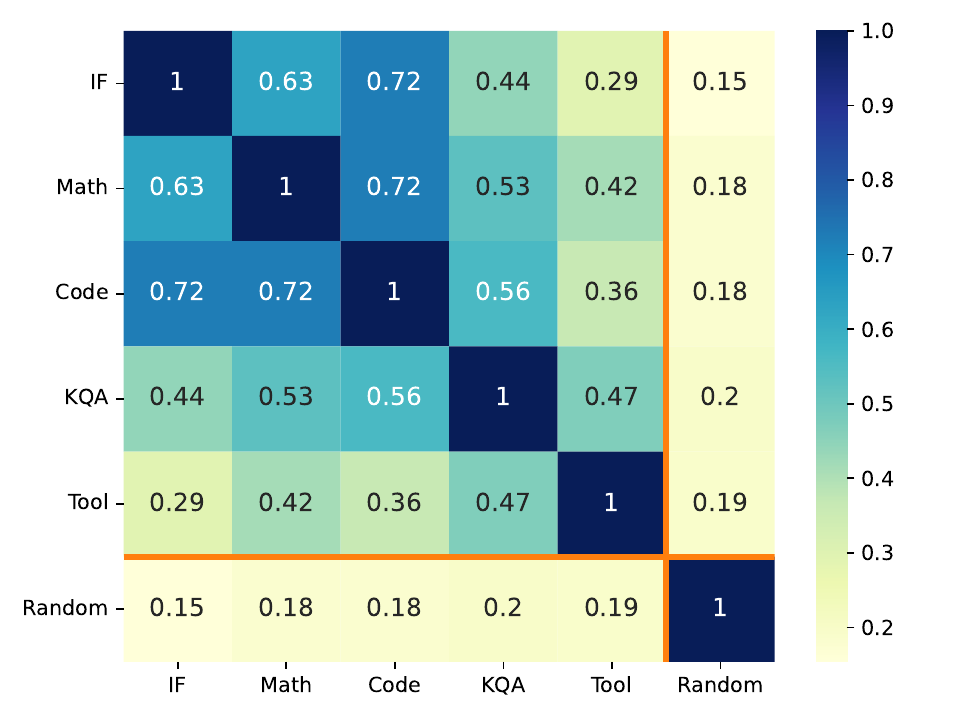}
        \caption*{moderate importance groups}

	\end{minipage}
    \begin{minipage}{0.32\linewidth}
		\centering
		\includegraphics[width=1\linewidth]{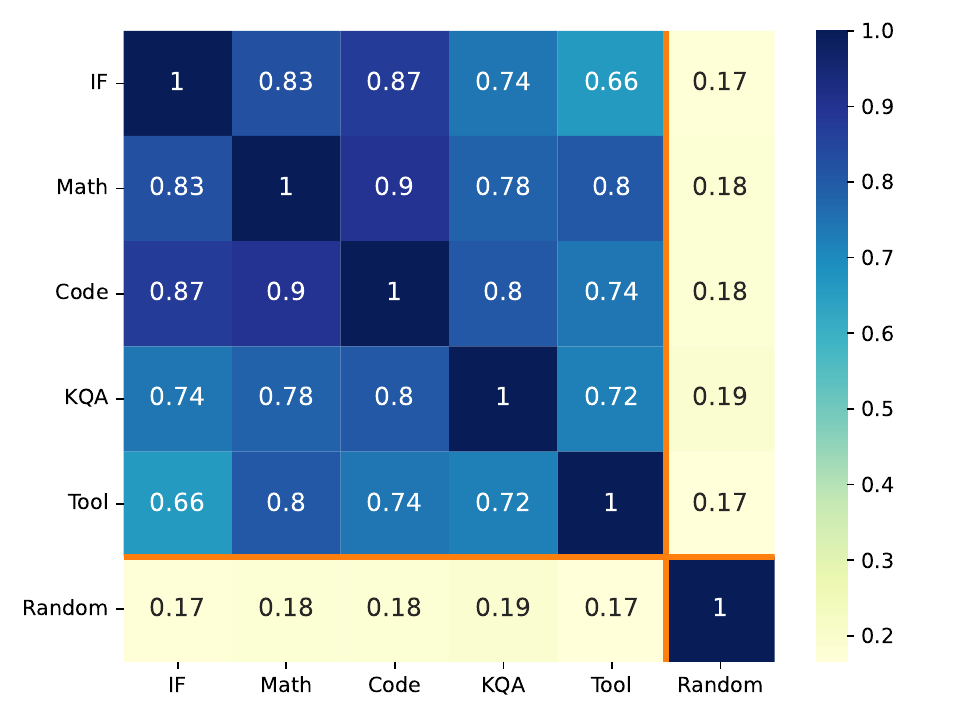}
        \caption*{low importance groups}

	\end{minipage}
    \caption{The Jaccard Index of different groups, \textit{Random} means sorting the components randomly.}
    \label{pic:Jaccard Index}
\end{figure*}

\subsection{Components Perspective}
\label{sec:analysis compontent}
\subsubsection{Analytical Approach}

Large language model, composing the decoder-base architecture of transformer, typically comprises multiple layers. 
Inspired by \citeauthor{bansal-etal-2023-rethinking} \shortcite{bansal-etal-2023-rethinking}, we adopt the gradient-based importance score to identify the important components.

In the auto-regressive language models, for the input $x$ and corresponding labels $y$ in given dataset $\mathcal{D}$, the loss function $\mathcal{L}$ to optimize the model is calculated as follows:
\begin{equation}
    \mathcal{L}(\mathcal{D}, \theta) = \sum_{(x,y)\in\mathcal{D}}\log p_\theta(y|x)
\end{equation}
where $\theta$ denotes the complete set of the model's parameters. For specific parameters set $\theta_{h}$, the importance score $\mathcal{I}_h$ is computed as:

\begin{equation}
\footnotesize
\begin{aligned}
\mathcal{I}_{h}(\mathcal{D}, \theta) &=\left|\mathcal{L}(\mathcal{D},\theta)-\mathcal{L}\left(\mathcal{D},\theta\mid\theta_{h}=0\right)\right| \\
& = \left|\frac {\partial \mathcal{L}}{\partial \theta_{h}} (\theta_{h} - 0) + \frac{1}{2!}\frac{\partial^{2} \mathcal{L}}{\partial \theta_{h}^{2}}(\theta_{h} - 0)^{ 2 } + \cdots \right| \\
& \approx \left| \theta_{h}^{T} \frac {\partial \mathcal{L} } { \partial \theta_{h} } \right| \\
\end{aligned}
\label{formula:importance_score}
\end{equation}

To accelerate the calculating speed, the score is estimated by calculating the first-order derivative of the Taylor expansion in the above formula.

\subsubsection{Experimental Results}

To examine the changes of components across various general tasks during the tool learning process, we select a subset of instructions from the training set as a proxy for a specific general ability $t$. Employing Equation~(\ref{formula:importance_score}), we determine the importance scores for the linear components in the vanilla model (Meta-Llama-3-8B-Instruct) with these data. In order to explore the effect of the importance of components, we conduct three experiments as follows:

\textbf{Experiment 1: }
We prioritize components based on their importance scores, and substitute the parameters of these components in the vanilla model with the corresponding weights from the model that has been fine-tuned on the tool learning dataset API-Bank, leaving all other parameters intact. We then assess the model's performance after replacement.

In Table~\ref{table:replacement on API-Bank}, we observe that Tool learning influences other abilities expressions in most cases. Following the replacement operation, the model's performance (GSM8K, HumanEval and MT-Bench) significantly decrease comparing to vanilla model, indicating that fine-tuning on the tool learning dataset stimulate tool-invoking abilities in these components while suppress their original inherent abilities. 
\textbf{More significant performance impacts occur when high-importance components are altered.} The performance of the model significantly drop after replacing weights with higher importance. Moreover, this phenomenon becomes more pronounced as the replacement ratio increases. 

\begin{table}[t]
\centering
{\small
    \setlength{\tabcolsep}{1mm}
	\begin{tabular}{c | c  c  c  c | c  } 
		\toprule
        \multirow{2}{*}{Dataset} & 
        \multicolumn{4}{c|}{Replace Ratio} &
        \multirow{2}{*}{Vanilla} 
        \\

        & T-20$\%$
        & T-50$\%$
        & D-20$\%$
        & D-50$\%$
        &
        
        \\
        \midrule
        GSM8K
        
        & 64.59
		& 40.56
        & 79.15
        & 74.98
        & 79.45
        
		\\
		HumanEval
        
        & 51.83
		& 25.00
        & 58.24
        & 20.12
        & 59.15
        
		\\
        TriviaQA
        
        & 64.95
		& 58.76
        & 64.95
        & 63.96
        & 64.82

		\\
        MT-Bench
        & 72.53
        & 50.97
		& 81.75
        & 73.19
        & 80.13
		\\
		
		\bottomrule
	\end{tabular}
}
	\caption{The performance of the model on the corresponding tasks after components parameter replacement. T-x$\%$ represents taking the top x$\%$ components of the model to be replaced by importance ranking order, D-x$\%$ is the opposite. Vanilla means test on Meta-Llama-3-8B-Instruct.}
	\label{table:replacement on API-Bank}
\end{table}

\textbf{Experiment 2: }
Furthermore, to delve into the inter-task relationships of components importance ranking, we apply Jaccard index to analyze the correlation between the importance ranking of model components. The Jaccard index can be represent as: $J(A, B) = |A \cap B|/|A \cup B|$, used to compare the differences and similarities between two samples $A$ and $B$.
Specifically, based on the importance ranking of components associated with ability $t$, we segment the components into three groups evenly: high, moderate and low. Then we compute the Jaccard Index between different abilities for each group respectively. For example, $J(C_h^t, C_h^m)$ represents the Jaccard Index between components groups of high importance calculated by ability $t$ and $m$ respectively.

As shown in Figure~\ref{pic:Jaccard Index}, \textbf{there is a high correlation across different tasks of their components importance ranking}. Certain components are pivotal in the expression of the model’s abilities across a broad range of tasks, whereas part of components exert a more limited influence on most tasks.

\textbf{Experiment 3: }
We try to inject the tool-invoking knowledge into the less important components as they store less tool-related ability. Considering a components set $H$ containing all linear module of the network, for tool learning dataset $\mathcal{D}_{tool}$, we compute the component level gradient-based importance scores for component $h$ is as follows:

\begin{equation}
\footnotesize
    \mathcal{M}_h(\theta) = \frac{\mathcal{I}_h(\mathcal{D}_{tool}, \theta)}{\sum_{i \in H}\mathcal{I}_i(\mathcal{D}_{tool}, \theta)}
\label{formula:M}
\end{equation}

In order to better combine the importance ranking of components and the training process, we select 20\% (Top, Down, Random) components in the model based on component importance $\mathcal{M}_h$ and train them using full parameter fine-tuning and LoRA methods, while keep other parameters frozen. Then we analyze the fine-tuning result detailed.

\begin{table}[htbp]
\centering
{\small
    \setlength{\tabcolsep}{2mm}
	\centering
	\begin{tabular}{l | c c | c  c  c  c } 
		\toprule
        
        \multirow{2}{*}{Model} 
        & \multicolumn{2}{c|}{Tool Ability}
        & \multicolumn{4}{c}{General Ability}
        \\

		& C
		& R
        & GSM
        & HE
        & TQA
        & MT
        \\
		\midrule
        \multicolumn{7}{c}{LoRA}
        \\
        \midrule

        Top 
		& 59.95
		& \textbf{40.85}
        & 74.00
        & 53.66
        & 64.09
        & \textbf{78.56}
        
        \\

        Down
        & 46.21
		& 38.84
        & \textbf{78.39}
        & \textbf{55.49}
        & 63.71
        & 74.00
        \\

    Random
        & \textbf{64.18}
		& 36.87
        & 75.36
        & 54.27
        & \textbf{65.11}
        & 76.40
        \\
        \midrule
        \multicolumn{7}{c}{Full parameter}
        \\
        \midrule

        Top 
		& \textbf{66.11} 
		& \textbf{44.04} 
        & 71.65 
        & 50.61 
        & \textbf{64.18} 
        & 73.15
        
        \\

        Down
        & 57.38 
		& 40.36 
        & \textbf{81.20}
        & \textbf{54.27}
        & 59.96
        & \textbf{74.11}
        \\

        Random
        & 59.31 
		& 37.99 
        & 74.30
        & 50.61
        & 63.77
        & 72.48
        \\

		\bottomrule
	\end{tabular}
}
	\caption{The fine-tuning experiments based on gradient-based importance score on the dataset API-Bank. The evaluation metrics C and R refer to Section~\ref{sec:evaluation_metrics}. 
	}
	\label{table:lora_allocation_and_unfreeze_module_result}
\end{table}

As illustrate in Table~\ref{table:lora_allocation_and_unfreeze_module_result}, we find that fine-tuning the components with lower importance score $\mathcal{M}_h$ have the least impact on the general performance, which confirms the existence of high correlations between different abilities of the models because $\mathcal{M}_h$ is computed by tool learning dataset. 
At the same time, it is observed that fine-tuning components with lower importance may not perform outstandingly in terms of tool-invoking ability. If only unimportant parameters were fine-tuned, the model can not thoroughly understand the format of tool callings\footnote{More analysis and details refer to Appendix~\ref{appendix:component_analysis}}. But ``Top'' and ``Random'' do not have such problems. This inspires us only fine-tuning the unimportant component sometimes may be not optimal.

\section{Methodology}
\begin{figure*}[htbp]
    \centering
    \includegraphics[width=0.9\textwidth]{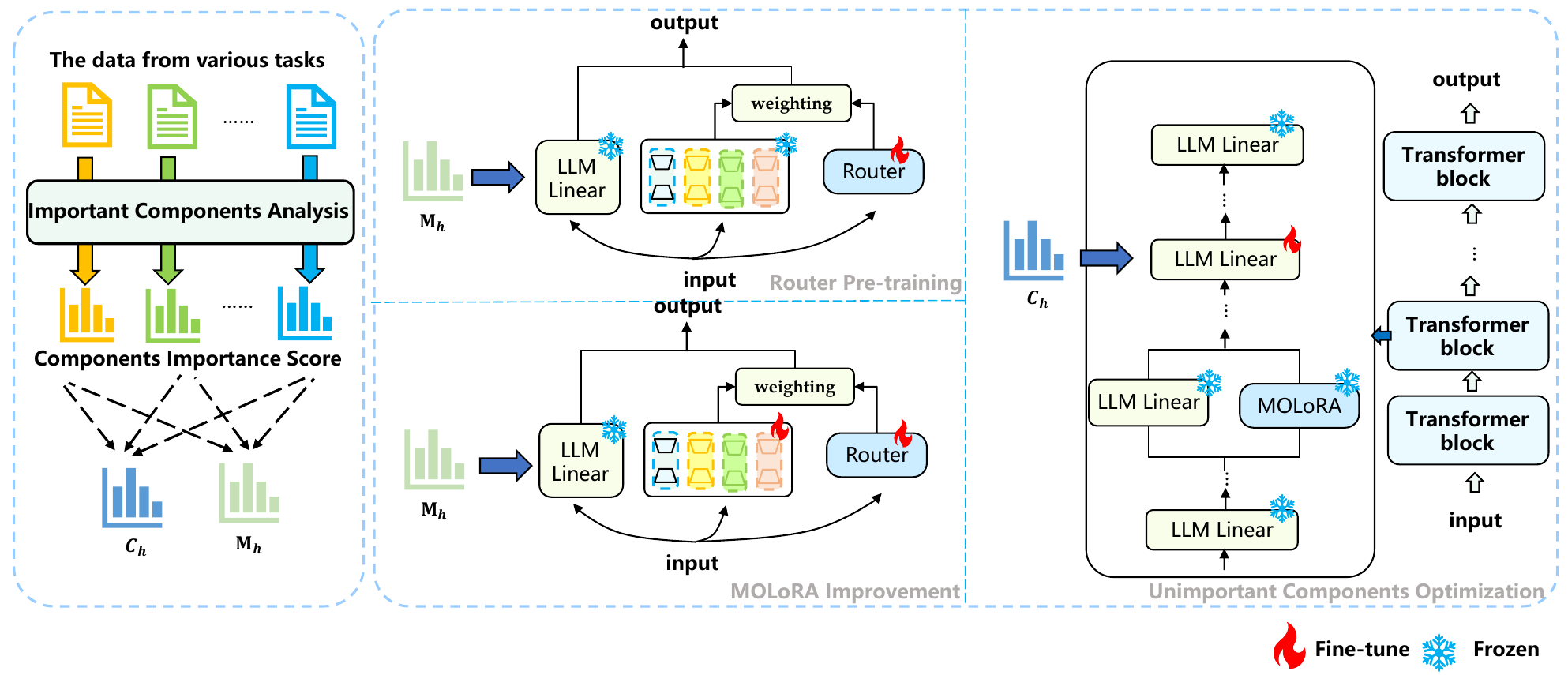}
    \caption{The overall architecture of \textbf{CITI}. The MOLoRA adapters are applied to important components identified by $\mathcal{M}_h$, and unimportant components identified by $\mathcal{C}_h$ are fine-tuned with all parameters.}
    \label{Figure:structure}
\end{figure*}

Instead of only focusing on the precise tool-callings, we propose a \textbf{C}omponent \textbf{I}mportance-based \textbf{T}ool-utilizing ability \textbf{I}njection method (\textbf{CITI}), injecting the tool-utilizing ability into the model while maintaining its general abilities. Based on the analysis experiments, in order to fully utilize the model's capability, we apply Mixture-of-LoRA adapters to important components and fine-tune a subset of unimportant components. As shown in Figure~\ref{Figure:structure}, our method mainly consists of three training stage: \textit{Router Pre-training} (RP), \textit{MOLoRA Improvement} (MI), \textit{Unimportant Components Optimization} (UCO). We will illustrate each technique in detail.

\subsection{Mixture of LoRA}
Based on the analysis of component importance, we want to fine-tune both important or unimportant components to achieve a comprehensive grasp of tool-utilizing knowledge. But adjusting critical components through fine-tuning can sometimes be detrimental to the overall network performance as the Co-directional shift phenomenon in hidden space. To alleviate this influence, a straightforward idea is to implement a gating function to distinguish tool-related and tool-unrelated inputs, and then assign different specialized experts to handle them separately.
To mitigate this, we implement MOLoRA adapters on these pivotal components with higher $\mathcal{M}_h$.

Inspired by \citeauthor{zhao2024matterenhancingemotionalintelligence} \shortcite{zhao2024matterenhancingemotionalintelligence}, we apply a router network with an additional neuron in MOLoRA to separate the inputs, allowing tool-related inputs are produced by combine of the LLM backbone and external LoRA experts. And the tool-unrelated data is predominantly processed by backbone itself. 

Meanwhile, mixing data during downstream task fine-tuning helps alleviate catastrophic forgetting of models and helps to avoid over-fitting. To train the router network and avoid catastrophic forgetting problems, we mix data from other fields during tool learning process.

For router network, we use Linear module as the gate function $G(x)$. Denote the input features as $x$, the formulas is as follows:

\begin{equation}
    G(x) = Softmax(W \cdot x)
\end{equation}
where $W \in \textsc{R}^{d \times (N+1)}$ is the trainable matrix, $d$ represents the dimensionality of the input, $N$ is the number of LoRA experts, The term $N+1$ means the incorporation of an additional neuron, representing the probability that $x$ is tool-unrelated data, which is used to suppress the model's call to external expert weights.
The output of MOLoRA is:
\begin{equation}
    h = W_0 x + \sum_{i=0}^{N} G(x)[i+1] B_i A_i x
\end{equation}
where $G(x)[i]$ is the $i$-th element in router output, and the $G(x)[0]$ is not used in forward process. And matrices $A\in \mathbb{R}^{r \times k}$ and $B\in \mathbb{R}^{d \times r}$ are the trainable linear of LoRA experts, where $r\ll min(d,k)$ is the rank of LoRA experts.

To achieve the separation of inputs, we divide the data into tool-related and unrelated with different task types. And we modify the Localized Balancing Constraint proposed by \citeauthor{dou2024loramoe} \shortcite{dou2024loramoe}, which softly constrains the experts to focus on different tasks. 
We assign the importance score to each neuron to control the weighting of different experts, the rules of importance matrix $I$ is:

\begin{equation}
    I_{n} = 
   \begin{cases} 
   [1 + \delta, \underbrace{1 - \delta, \cdots , 1 - \delta}_N] & \text{if } x_n \in \mathcal{D}_{tool}  \\
   [1 - \delta, \underbrace{1 + \delta, \cdots , 1 + \delta}_N] & \text{otherwise}
   \end{cases}
\end{equation}
   
Here, $\delta \in [0, 1]$ controls the degree of imbalance between router output, $x_n$ is the $n$-th input.

We weight the importance matrix $I$ and the output of the gating function $G$, denoted as $Z = I \odot G$. And the routing loss $\mathcal{L}_{r}$ of router network is:
\begin{equation}
    \mathcal{L}_{r} = \frac{\sigma^2(Z)}{\mu(Z)}
\end{equation}
where $\sigma^2(Z)$ and $\mu(Z)$ are the variance and mean of $Z$. $\mathcal{L}_{r}$ ensure that for tool-unrelated data $x$, $G(x)[0]$ is relatively large, and vice versa, $G(x)[0]$ is relatively small. adjusting the weight of LoRA experts based on the data types.

\subsection{Unimportant Components Optimization}

For unimportant components, as updating these parameters has a relatively less impact on the model's general capabilities, we utilize full parameter fine-tuning to stimulate the model's tool invoking ability.

We select unimportant components for general abilities by averaging the importance of different tasks, defined as general abilities importance $\mathcal{C}_h(\theta)$. 
For a general abilities set $T$, the formula is as follows:

\begin{equation}
    \mathcal{C}_h(\theta) = \sum_{t \in T} \frac{\mathcal{I}_h(\mathcal{D}_t, \theta)}{\sum_{i \in H}\mathcal{I}_i(\mathcal{D}_t, \theta)} 
\label{formula:C}
\end{equation}

Components with lower $\mathcal{C}_h(\theta)$ are selected. These components are not frozen during the training process.

\subsection{Training Strategy}
In summary, the overall loss function of \textbf{CITI} is:
\begin{equation}
\footnotesize
\mathcal{L}= -\sum_{(x,y)\in\mathcal{D}}\log P\left(y\mid x; \theta_b,\theta_e,\theta_r\right) + \beta \cdot \mathcal{L}_{r}
\end{equation}
Where $\theta_b$ is the parameters of the backbone, $\theta_e$ represents the MOLoRA experts and $\theta_r$ is the router network. $\mathcal{D}$ denotes the mixture of tool learning dataset and other instructions sampled from general ability datasets.

To optimize our model effectively, we have devised a three-stage training strategy, delineated as follows:

\textbf{Stage-1 RP}: We integrate MOLoRA based on the importance of component gradients. During this phase, the core model parameters and LoRA experts are kept constant, with only $\theta_r$ being subjected to training.

\textbf{Stage-2 MI}: The $\theta_e$ and $\theta_r$ are subjected to training. We fine-tune the parameters in LoRA experts.

\textbf{Stage-3 UCO}: Here, we immobilize the parameters of the MOLoRA adapters, including $\theta_e$ and $\theta_r$, as well as the important parameters in backbone identified by importance ranking. We then proceed to train a select subset of the backbone parameters chosen by $\mathcal{C}_h$.

\section{Experiments}
\subsection{Datasets}
We conduct the experiments on two tool learning benchmarks: API-Bank \cite{li-etal-2023-api} and ToolAlpaca \cite{tang2023toolalpaca}. 
Additionally, we assess general abilities through experiments on datasets mentioned above.

\subsection{Implementation Details}
We apply MOLoRA to top 20\% components with highest importance ranking sorted by $\mathcal{M}_h$, and fine-tune down 10\% components with lowest importance ranking sorted by $\mathcal{C}_h$. More training details refer to Appendix~\ref{appendix:implementation}.

\subsection{Evaluation Metrics}
\label{sec:evaluation_metrics}
For API-Bank dataset, we follow the evaluation metrics propose by \citeauthor{li-etal-2023-api}\shortcite{li-etal-2023-api}, including the \textbf{C}orrectness of API calls and the \textbf{R}OUGE-L to test quality of the responses. 
For ToolAlpaca dataset, we utilize the GPT-4 to evaluate the tool-invoking process in real-world testing subset and follow the original metrics: \textbf{P}rocedure: The correctness of tool utilizing procedure, \textbf{R}esponse: The quality of final response, \textbf{O}verall: Whether procedure and response are both correct.

\subsection{Baselines}
We conduct experiments with excellent models: \textbf{Meta-Llama-3-8B-Instruct}, \textbf{Phi-3-mini-128k-instruct} and \textbf{Mistral-7B-Instruct-v0.2}. We compare our method with the following baselines:
(1) \textbf{FT}: fine-tune the model with full parameter on the tool learning dataset; (2) \textbf{LoRA}: fine-tune the model with LoRA adapters on tool learning dataset. 
The training data is without data mixing for our baselines.

\begin{table}[t]
\centering
\setlength{\tabcolsep}{2mm}
{\small

	\centering
	\begin{tabular}{l | c c  | c  c  c  c } %
		\toprule

        \multirow{2}{*}{Model} 
        & \multicolumn{2}{c|}{Tool Ability}
        & \multicolumn{4}{c}{General Ability}
        \\

		& C
		& R
        & GSM
        & HE
        & TQA
        & MT
        \\
		\midrule

        Llama-3
		& 33.25 
		& 14.67 
        & 79.45
        & 59.15
        & 64.82
        & 80.13
        
		\\
  
		FT
		& 56.48 
        & 18.71 
        & 26.38
        & 0.00
        & 50.07
        & 33.88 

        \\

        LoRA
		& \textbf{65.98} 
        & \textbf{38.62} 
        & 76.88
        & 53.05
        & \textbf{64.84}
        & 68.00

        \\

        \midrule

        \textbf{CITI}
		& 58.92 
		& 34.51 
        & \textbf{77.33} 
        & \textbf{54.88} 
        & 64.69 
        & \textbf{76.94}
        
        \\

        \midrule
        
        Phi-3
		& 44.80 
		& 29.40 
        & 82.71
        & 60.37
        & 53.08
        & 81.28
        
		\\
  
		FT
        & \textbf{61.75}
		& 35.23 
        & 73.09
        & 51.83
        & 53.32
        & 66.81

        \\

        LoRA
        & 57.00 
		& 35.19 
        & 75.82
        & 0.00
        & 52.34
        & 68.44

        \\

        \midrule

        \textbf{CITI} 
        & 58.02 
		& \textbf{37.21} 
        & \textbf{77.86} 
        & \textbf{61.59}
        & \textbf{55.23} 
        & \textbf{72.19}
        
        \\
        \midrule

        Mistral
		& 53.66
		& 32.39
        & 44.81
        & 35.37
        & 59.21
        & 75.66
        
		\\
  
		FT
        & 47.88
		& 2.18
        & 2.05
        & 0.00
        & 13.50
        & 10.63

        \\
        
        LoRA
        & 58.66
		& \textbf{43.31}
        & 46.32
        & 37.80
        & 61.54
        & 68.31 

        \\

        \midrule

        \textbf{CITI} 
        & \textbf{62.00}
		& 33.53
        & \textbf{53.30}
        & \textbf{38.41}
        & \textbf{62.77}
        & \textbf{69.28}
        
        \\

		\bottomrule
	\end{tabular}
}
	\caption{The overall results on the dataset API-Bank.
	}
	\label{table:apibank_result}
\end{table}

\begin{table}[t]
\centering
{\small
\setlength{\tabcolsep}{1.2mm}
	\centering
	\begin{tabular}{l | c c c | c  c  c  c  } 
		\toprule

        \multirow{2}{*}{Model} 
        & \multicolumn{3}{c|}{Tool Ability}
        & \multicolumn{4}{c}{General Ability}
        \\

		& P
		& R
        & O
        & GSM
        & HE
        & TQA
        & MT
        \\
		\midrule
        
        Llama-3
		& 14.91
		& 21.05
        & 12.28
        & 79.45
        & 59.15
        & 64.82
        & 80.13
        
		\\
  
		FT
		& 65.79
		& 59.65
        & 55.26
        & 30.33
        & 0.61
        & 55.97
        & 50.81

        \\

        LoRA
		& \textbf{76.32}
		& 74.56
        & 71.05
        & 73.46 
        & 31.71 
        & \textbf{65.18} 
        & 73.75

        \\

        \midrule

        \textbf{CITI} 
		& \textbf{76.32}
		& \textbf{78.07}
        & \textbf{72.81}
        & \textbf{75.89}
        & \textbf{56.71}
        & 64.82
        & \textbf{76.67}
        
        \\

        \midrule
        
        Phi-3
		& 2.63
		& 1.75
        & 1.75
        & 82.71
        & 60.37
        & 53.08
        & 81.28
        
		\\
  
		FT
        & 67.54
		& 68.42
        & 64.04
        & 71.95
        & 54.88
        & 53.58
        & 65.06 

        \\

        LoRA
        & \textbf{70.18}
		& 71.05
        & \textbf{66.67}
        & 74.91
        & 0.61
        & 53.04
        & \textbf{73.22} 

        \\

        \midrule

        \textbf{CITI}
        & 64.04
		& \textbf{72.81}
        & 61.40
        & \textbf{77.33}
        & \textbf{62.20}
        & \textbf{55.27}
        & 72.59 
        
        \\
        \midrule

        Mistral
		& 14.04
		& 14.91
        & 13.16
        & 44.81
        & 35.37
        & 59.21
        & 75.66
        
		\\
  
		FT
        & 42.98
		& 41.23
        & 30.70
        & 4.85
        & 0.00
        & 12.10
        & 15.69 

        \\
        
        LoRA
        & 72.81
		& 73.68
        & 67.54
        & 44.81
        & 34.76
        & 62.99
        & 68.19 

        \\

        \midrule

        \textbf{CITI}
        & \textbf{78.07}
		& \textbf{76.32}
        & \textbf{71.05}
        & \textbf{51.48}
        & \textbf{36.59}
        & \textbf{63.21}
        & \textbf{71.44}
        
        \\

		\bottomrule
	\end{tabular}
}
	\caption{The overall results on the dataset ToolAlpaca.
	}
	\label{table:toolalpaca_result}
\end{table}

\subsection{Overall Results}

As shown in Table~\ref{table:apibank_result} and Table~\ref{table:toolalpaca_result}, CITI demonstrates effective performance in preserving general abilities in two datasets. For instance, it achieves 7.59\% improvement comparing to LoRA and 31.95\% comparing to FT in API-Bank for general abilities in average. 
For tool-utilizing ability, CITI shows competitive results in two datasets compared to the baselines. Noticing that in the dataset API-Bank, CITI has poorer tool-invoking ability compared to the baselines (e.g. ROUGE-L score in Mistral). We suggest there are two possible reasons: 
(1) The $\lambda$ in MOLoRA may not be optimal. And insufficient training data may result in underfitting or overfitting of LoRA adapters.
(2) The evaluation metrics are not comprehensive. API-Bank asks model to generate tool-calling or response based on conversation history, but the path for model to obtain final correct answers is not unique.

Additionally, we notice that LoRA outperforms full parameter fine-tuning in both tool utilization and general abilities in most cases. This may because full parameter fine-tuning can easily lead to overfitting.

\begin{table}[htbp]
\centering
\setlength{\tabcolsep}{1.5mm}
{\small
	\centering
	\begin{tabular}{l | c c | c  c  c  c }
		\toprule
        
        \multirow{2}{*}{Model} 
        & \multicolumn{2}{c|}{Tool Ability}
        & \multicolumn{4}{c}{General Ability}

        \\

		& C
		& R
        & GSK
        & HE
        & TQA
        & MT
        \\
		\midrule

        FT + DM
		& 56.61
		& 39.67
        & 54.44
        & 29.88
        & 45.28
        & 60.41
        
        \\
        LoRA + DM 
		& \textbf{63.03}
		& 38.66
        & 75.13
        & 51.22
        & 63.96
        & 74.31
        
        \\
        \midrule

        UCO
        & 59.31
		& 38.63
        & \textbf{79.68}
        & 54.88
        & 64.07
        & 75.09
        \\

        RP + MI
        & 57.89 
		& 35.50 
        & 77.94 
        & 54.88 
        & \textbf{64.74} 
        & 76.56
        \\
        \ \ w/o MOLoRA
        & 61.87 
		& \textbf{39.97} 
        & 73.09 
        & 54.27 
        & 63.09 
        & \textbf{77.91}
        \\
        \ \ w/o $\mathcal{L}_{r}$
        & 61.49
		& 38.29
        & 75.28
		& \textbf{57.93}
        & 63.90
        & 76.81
        \\

        \textbf{CITI} (ours) 
		& 58.92 
		& 34.51 
        & 77.33 
        & 54.88 
        & 64.69 
        & 76.94
        
        \\
        \ \ w/o \textbf{RP}
        & 58.79
		& 38.13
        & 76.12
        & 52.44
        & 63.35
        & 75.22
        \\

		\bottomrule
	\end{tabular}
}
	\caption{The 
 ablation results on the dataset API-Bank.
	}
	\label{table:apibank_ablation_result}
\end{table}

\begin{table}[htbp]
\centering
\setlength{\tabcolsep}{1mm}
{\small
    
	\centering
	\begin{tabular}{l | c c c | c  c  c  c } 
		\toprule
        
        \multirow{2}{*}{Model} 
        & \multicolumn{3}{c|}{Tool Ability}
        & \multicolumn{4}{c}{General Ability}
        \\

		& P
		& R
        & O
        & GSM
        & HE
        & TQA
        & MT
        \\
		\midrule

        FT + DM
		& 60.53
		& 63.16
        & 57.02
        & 52.92
        & 23.78
        & 50.58
        & 58.94
        
        \\
        LoRA + DM 
		& 73.68
		& 73.68
        & 68.42
        & 74.53
        & 53.66
        & \textbf{65.06}
        & 74.94
        
        \\
        \midrule

        UCO
        & 69.30
		& 68.42
        & 64.04
        & \textbf{80.52}
        & 55.49
        & 64.06 
        & 73.63 
        \\
        RP + MI
        & 76.32
		& 75.44
        & 69.30
        & 74.68
        & 54.27
        & 64.49
        & \textbf{77.63}
        \\
        \ \ w/o MOLoRA
        & 70.18
		& 71.05
        & 65.79
        & 73.09
        & 55.49
        & 64.31
        & 77.31
        \\
        \ \ w/o $\mathcal{L}_{r}$
        & 71.05
		& 71.93
        & 64.91
		& 77.48
        & 55.49
        & 65.00
        & 77.38
        \\

        \textbf{CITI} (ours) 
		& \textbf{76.32}
		& \textbf{78.07}
        & \textbf{72.81}
        & 75.89
        & \textbf{56.71}
        & 64.82
        & 76.67
        
        \\
        
        \ \ w/o RP
        & 70.18
		& 68.42
        & 64.91
        & 77.48
        & 54.27
        & 64.95
        & 76.81
        \\

		\bottomrule
	\end{tabular}
}
	\caption{The 
 ablation results on the dataset ToolAlpaca.
	}
	\label{table:toolalpaca_ablation_result}
\end{table}

\subsection{Ablation Studies}
To demonstrate the effectiveness of our approach, we conduct ablation studies on model Meta-Llama-3-8B-Instruct as shown in Table~\ref{table:apibank_ablation_result} and Table~\ref{table:toolalpaca_ablation_result}. Where w/o MOLoRA means replacing the MOLoRA by LoRA adapter, w/o $\mathcal{L}_r$ means applying normal MOLoRA without additional neuron and importance matrix $I$ in router. DM represents data mixing in the training data. The experimental results show that comparing to FT and LoRA with data mixing, CITI still exhibits advantages in general abilities. Additionally, we analyze the contributions of each module in CITI. Compared to baselines, MI and UCO modules have shown distinct advantages in maintaining model's general abilities separately.

We can find that:
(1) UCO module demonstrates that fine-tuning the unimportant components selectively can enhance the model's capability to utilize new tools while maintaining its original performance, especially on GSM8K within two datasets. 
(2) Comparing RP + MI with w/o MOLoRA, we discover that by incorporating MOLoRA, rather than simply adding LoRA adapter, the decline in general abilities can be alleviated. 
(3) The results of only w/o $\mathcal{L}_r$ outperform w/o MOLoRA in most cases, further indicting MOLoRA structure is effective. Additionally, we find that some results in w/o $\mathcal{L}_r$ are better than RP + MI. We deduce this is because the adapters absorb general knowledge within the general instructions in mixed training set, but they play little impact in MI because the router separates them into frozen LLM backbone during training stage.
(4) The results of w/o RP demonstrate that \textit{Router Pre-training} offers benefits especially in tool-utilizing in ToolAlpaca and general performance in API-Bank, indicating that pre-training the router network may accelerate model convergence especially when training data is constrained. 
(5) By applying UCF to further fine-tune the model following MI, CITI can maintain or even increase its general abilities on two datasets comparing to RP + MI. And the tool-utilizing ability has further significantly improved on the dataset ToolAlpaca.

\section{Related Work}
\subsection{Tool Learning}
Tool Learning aims to enable LLMs to use external tools to enhance models' ability, which shows great power in dealing with various tasks \cite{qin2023tool, qu2024tool}. Fine-tuning technology \cite{tang2023toolalpaca, qin2024toolllm, wang2024llms} is a primary approach to learn tool-utilizing pattern, which can solidify knowledge into models comparing to in-context learning \cite{shen2023hugginggpt, wu2023visual}.  
In order to enable model to invoke tools while maintaining its original performance, we introduce \textbf{CITI}, a framework to balance the conflict.

\subsection{Components Importance Analysis}
It is important to evaluate the importance of parameters in model pruning \cite{pmlr-v162-zhang22ao} and works related to interpretability \cite{NEURIPS2019_2c601ad9, zhang2024unveiling, jin2024cutting}. The gradient-based importance scores \cite{bansal-etal-2023-rethinking} assesses the importance of parameters by measuring the resultant impact on the loss function following the parameter's elimination. In our work,  We use importance scores to identify and select components on tool-utilizing and general tasks.

\subsection{Mixture Of LoRA}
Recently, there are substantial works utilizing Mixture-of-LoRA Experts (MOLoRA) architecture to improve the performance \cite{dou2024loramoe,wu2024mixture, yang2024moralmoeaugmentedlora}.
These works combine the Mixture-of-Experts (MOE) \cite{10.5555/3586589.3586709, shazeer2017} architecture with Low-Rank Adaptation (LoRA) \cite{hu2022lora, wu2024mixture, gou2024mixture}. MOLoRA keeps the backbone networks frozen and fine-tunes the adapters, achieving significant performance while alleviate catastrophic forgetting \cite{zhao2024matterenhancingemotionalintelligence, li2024moectnovelapproachlarge, dou2024loramoe}. We modify the router network to distinguish the tool-related and tool-unrelated inputs.

\section{Conclusion}
In this study, we explore the trade-offs between model's tool-utilizing ability and its general task performance following fine-tuning on a tool learning dataset. We conduct a deep analysis of this phenomenon from hidden representation and components perspectives and propose a new approach \textbf{CITI}, which integrates MOLoRA adapters into the critical components of the model and selectively fine-tunes a limited number parameters of less critical components within the backbone of the model. Through extensive experiments, we demonstrate the validity of \textbf{CITI} in acquiring tool-related knowledge and without excessively harming its performance on general tasks.

\bibliography{aaai25}
\newpage
\appendix
\section{Implementation Details}
\label{appendix:implementation}
\subsection{Importance Score Calculating}
We calculate the gradient-base importance of the components in LLMs. In our experiments, we calculate the importance of components in different tasks, which represents one ability of LLM. The tasks are as follows: coding, mathematics, factual knowledge, instruction following and tool-utilizing. The datasets used to compute importance score are same as the datasets mentioned in training set. For each dataset, we randomly sample 3000 examples to compute the importance score. 

To calculate the importance scores $\mathcal{C}_h$, 
in our experiments, $T$ contains mathematics, code generation, factual knowledge and instruction following abilities.

The code for importance computing is modified from llm-interpret\footnote{https://github.com/amazon-science/llm-interpret}. 

\subsection{Training Set Construction}
To train \textbf{CITI}, We random sample 5000 instructions from datasets in the field of coding (CodeAlpaca-20K)\footnote{https://huggingface.co/datasets/sahil2801/CodeAlpaca-20k}, mathematics (MetaMathQA)\footnote{https://huggingface.co/datasets/meta-math/MetaMathQA}, factual knowledge (TriviaQA)\footnote{https://huggingface.co/datasets/mandarjoshi/trivia\_qa} and instruction following\footnote{https://github.com/Instruction-Tuning-with-GPT-4/GPT-4-LLM} respectively, and mix them with the original tool training data in API-Bank. For ToolAlpaca, we sample 3000 training data from each datasets respectively.

\subsection{Training Details}
Our models are trained by LLaMA-Factory\footnote{https://github.com/hiyouga/LLaMA-Factory}. And we modify the dialogue template of training data to make it fit the template required by different models. All of experiments are conducted on NVIDIA A100 and RTX 3090 GPUs.
In \textbf{CITI}, each MOLoRA adapters contains 4 LoRA experts with rank value of 8 and alpha value of 2. We apply MOLoRA to top 20\% components with highest importance ranking sorted by $\mathcal{M}_h$, and fine-tune down 10\% components with lowest importance ranking sorted by $\mathcal{C}_h$. We train the model for 1 epoch with learning rate 2e-4 in stage \textit{Router Pre-training}, 2 epoch with learning rate of 2e-4 in stage \textit{MOLoRA Improvement}, 1 epoch with learning rate of 2e-5 in stage \textit{Unimportant Components Optimization}. 
The hyperparameters $\lambda$ is set to 0.95 and $\beta$ is set to 0.01.

For baselines in overall results, all of the model is trained for two epochs without data mixing. The learning rate for FT is 2e-5 with the batch size of 64, while that for LoRA is 2e-4 with the batch size of 64. In LoRA, all of the linear modules are applied LoRA adapters, the rank of adapters is 16 and alpha value is 16.

Meanwhile, for ablation experiments, FT + DM and LoRA + DM use the same training data as \textbf{CITI}. And w/o MOLoRA applies LoRA adapters with the rank value 32 and alpha value 32 to top 20\% components, which is similar to \textbf{CITI}. 

\subsection{Evaluation}
The evaluation of general abilities follow the setting from OpenCompass\footnote{https://github.com/open-compass/opencompass} and MT-Bench\footnote{https://github.com/lm-sys/FastChat/tree/main/fastchat/llm\_judge}. For fair comparison, the evaluation results of MT-Bench will be converted into a percentage scale in this passage. For evaluation of HumanEval, it matches the code content through regular expressions and then evaluates it, but after the FT and LoRA, the model may generate code content that does not adhere to the regular expression format, which cause the expression to not match the code, affecting the result. But in order to conduct a unified evaluation, we use the identical settings in OpenCompass to evaluate it.

In the testing process, we test the model's tool utilizing ability using the respective test sets from \textit{API-Bank} and \textit{ToolAlpaca}, we rewrite the dialogue template of the testing data and part of prompt to make it suitable for different LLMs.

\subsection{Details of Components Analysis}
\label{appendix:component_analysis}
The details of experiment 3 about the analysis of components are as follows:

For experiments 3, the rank of the LoRA is 32 and alpha is 32, the model is trained for 2 epochs. And Full parameters fine-tuning is trained for 1 epoch. The training data is the same as \textbf{CITI} with data mixing.

Besides, we analyze the predicting result in level-1-API of different models with different unfrozen components determined by importance score $\mathcal{M}_h$. level-1-API contains 399 testing instances. The statistical data is shown in Table~\ref{table:prediction_analysis}. The results suggest that, no matter LoRA or full parameter fine-tuning, only fine-tuning the ``Down'' importance components tend to error ``NO\_API\_CALL'' in predicting. But ``Top'' and ``Random'' don't have such problem.

\begin{figure}[htbp]
  \includegraphics[width=\columnwidth]{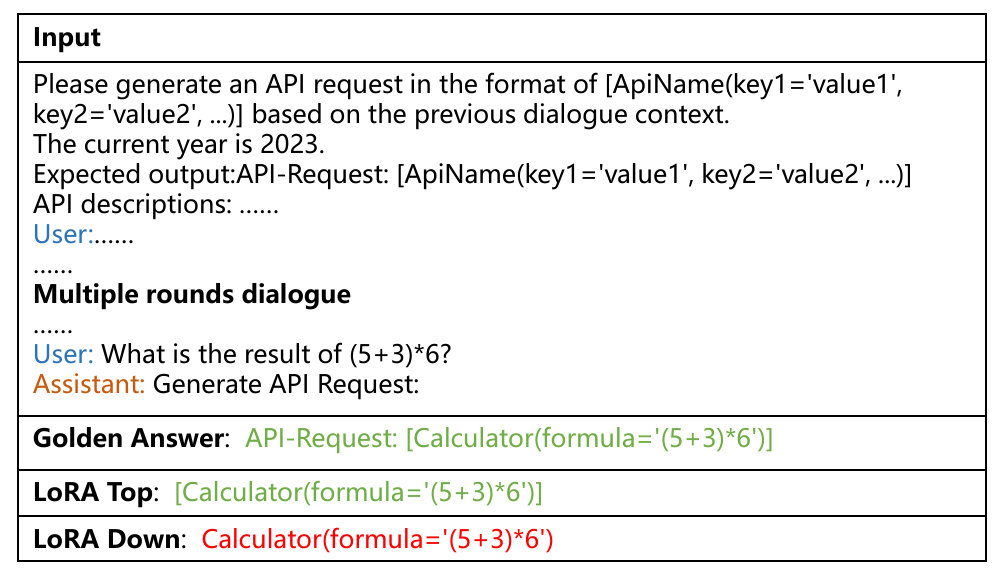}
  \caption{Case study of model's output}
  \label{fig:casestudy}
\end{figure}

In Figure~\ref{fig:casestudy}, there is the case study of the ``NO\_API\_CALL''. We think these phenomenons suggest that by only training the low importance components, the model may not fully understand the correct format to invoke tools. So fine-tuning both important and unimportant components may contribute to accurate tool-utilizing.

\begin{table*}[htbp]
\centering
{
    
	\centering
	\begin{tabular}{l | c c c  c  c  c  } 
		\toprule
        
        Error Type
        & LoRA Down
        & LoRA Random
        & LoRA Top
        & FT Down
        & FT Random
        & FT Top
        
        \\
        \midrule

		NO\_API\_CALL
		& 183
        & 0
        & 0
        & 60
        & 0
        & 0
        \\

		API\_NAME\_MISMATCH
		& 7
        & 18
        & 9
        & 17
        & 12
        & 8
        
        \\
        HAS\_EXCEPTION
		& 31
        & 43
        & 58
        & 47
        & 58
        & 51
        
        \\

        INPUT\_MISMATCH
        & 25
        & 29
        & 43
        & 38
        & 40
        & 39
        \\
        OUTPUT\_MISMATCH
        & 1
        & 7
        & 7
        & 2
        & 7
        & 5
        \\
        INVALID\_INPUT\_PARAMETER
        & 0
        & 0
        & 1
        & 0
        & 1
        & 0
        \\
        
        KEY\_ERROR
        & 2
        & 11
        & 11
        & 4
        & 11
        & 11
        \\

        MISS\_INPUT\_ARGUMENT
        & 0
        & 0
        & 1
        & 4
        & 11
        & 1
        \\

        Correct\_API\_calls
        & 150
        & 292
        & 270
        & 231
        & 271
        & 185
        \\

		\bottomrule
	\end{tabular}
}
	\caption{The analysis of prediction of testing data level-1-API in the \textit{API-Bank} dataset.
	}
	\label{table:prediction_analysis}
\end{table*}

\begin{figure*}[htbp]
  \includegraphics[width=1\textwidth]{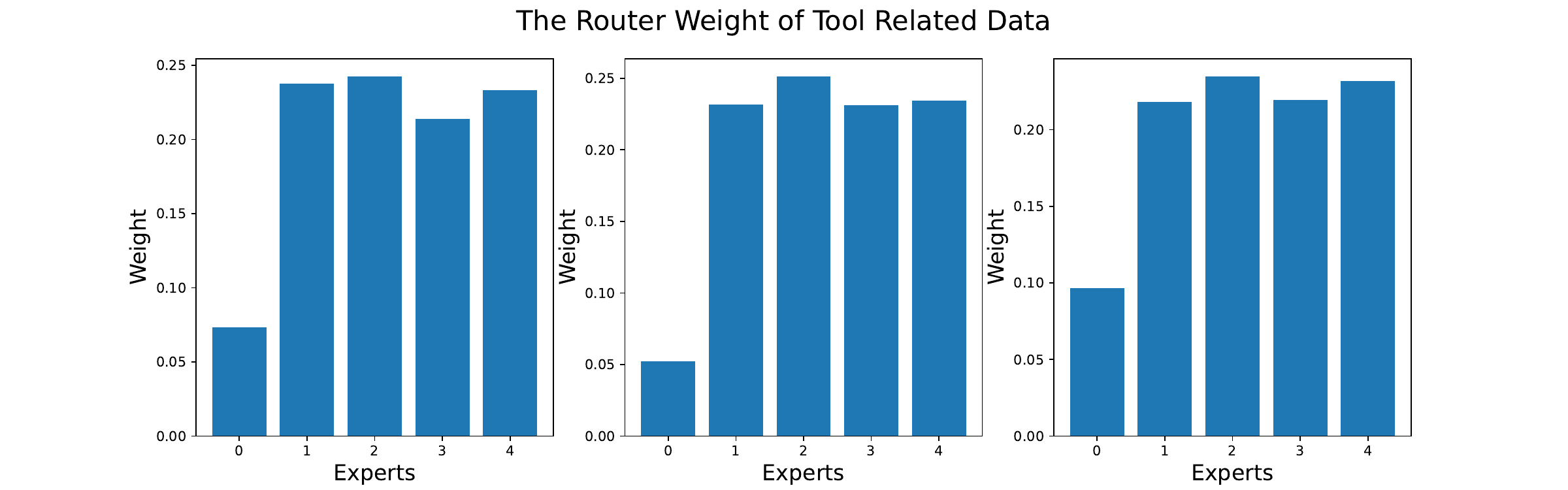}
  \caption{Router weight allocation of tool-related data}
  \label{fig:tool_weight}
\end{figure*}

\begin{figure*}[htbp]
  \includegraphics[width=1\textwidth]{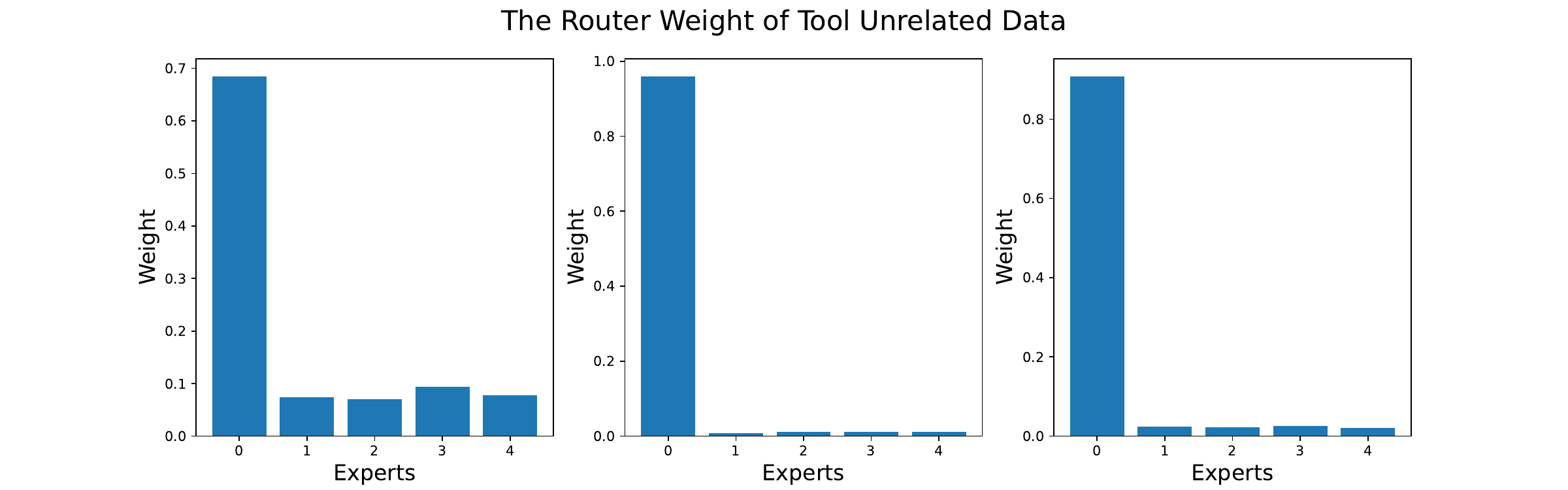}
  \caption{Router weight allocation of tool-unrelated data}
  \label{fig:untool_weight}
\end{figure*}

\section{Router Weight Visualization}
We selected several MOLoRA routers from the model and selected 100 tool-related and tool-unrelated data from the test data, collecting the output of the router network for visualization. The results are shown in Figure~\ref{fig:tool_weight} and Figure~\ref{fig:untool_weight}

\begin{figure*}[htbp]
	\centering
	\begin{minipage}{0.49\linewidth}
		\centering
		\includegraphics[width=1\linewidth]{mlp.pdf}
		\caption*{Feed-Forward Network}
	\end{minipage}
    \begin{minipage}{0.49\linewidth}
		\centering
		\includegraphics[width=1\linewidth]{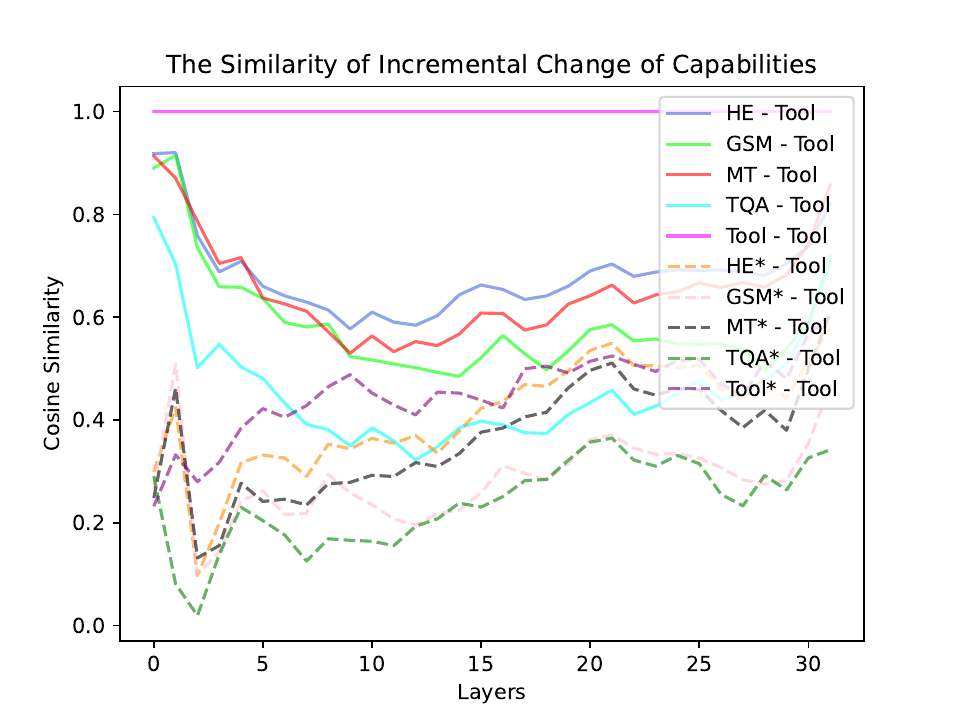}
        \caption*{Multi-Head Attention}

	\end{minipage}

    \caption{Cosine similarity of $ICC$ between the input of different layers of module in model Meta-Llama-3-8B-Instruct calculated by FT, where the notation with an asterisk (*) represents $ICC$ fine-tuned on the code-related dataset (e.g. TQA* represents $ICC$ of \textit{TriviaQA} trained by code dataset), and no asterisk (*) represents fine-tuned on tool learning dataset.}
    \label{pic:Sim}
\end{figure*}

\begin{figure*}[htbp]
	\centering
	\begin{minipage}{0.49\linewidth}
		\centering
		\includegraphics[width=1\linewidth]{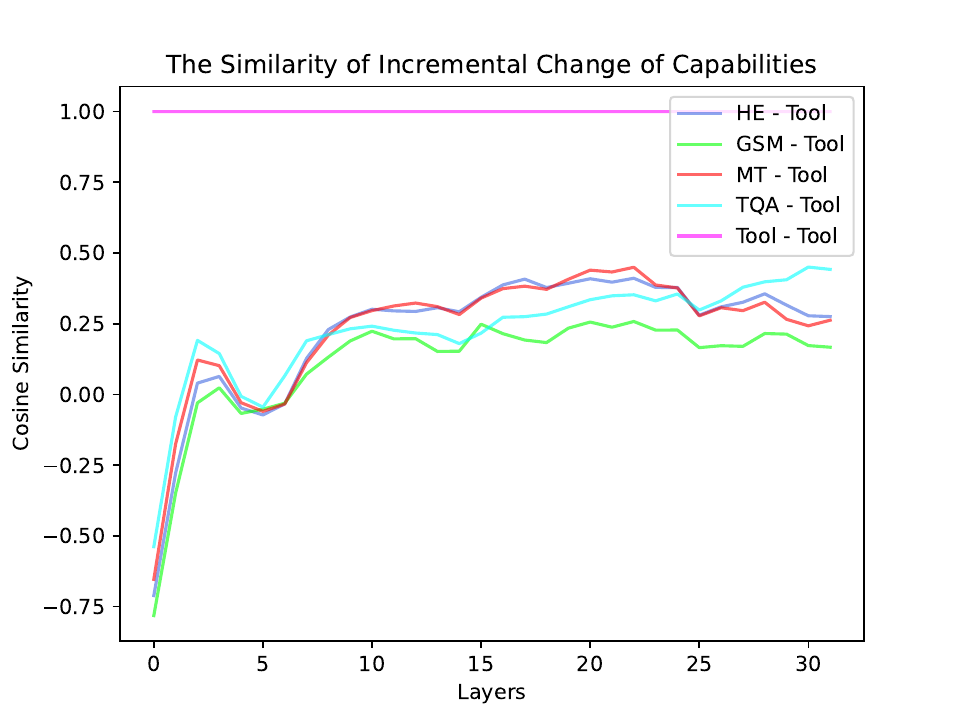}
		\caption*{Feed-Forward Network}
	\end{minipage}
    \begin{minipage}{0.49\linewidth}
		\centering
		\includegraphics[width=1\linewidth]{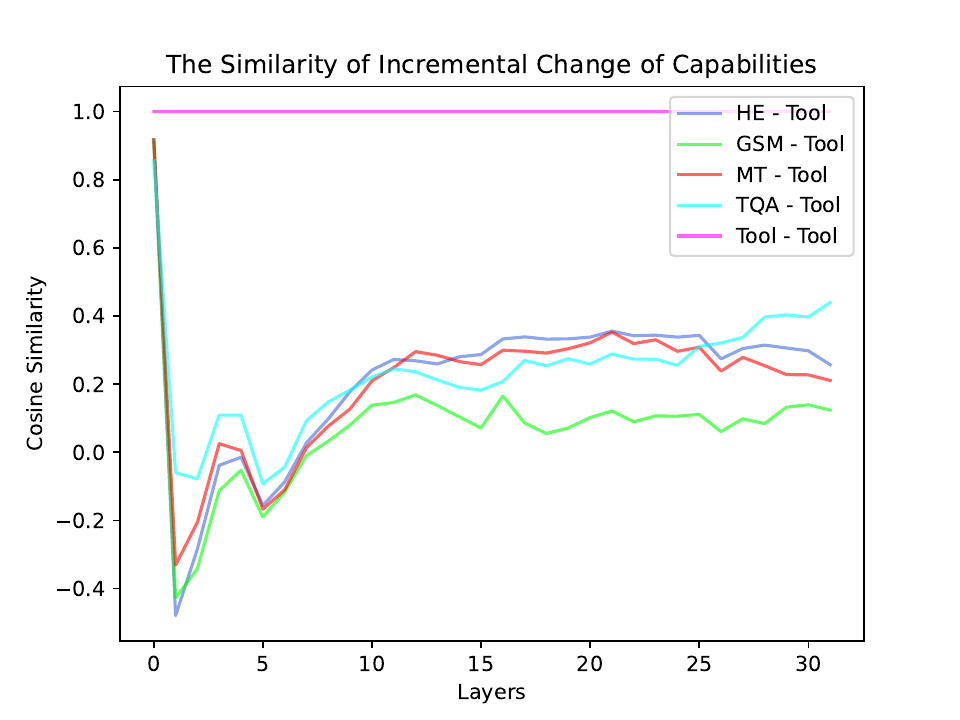}
        \caption*{Attention}
	\end{minipage}

    \caption{Cosine similarity of $ICC$ between the input of different layers of module in model Meta-Llama-3-8B-Instruct calculated by \textbf{CITI}. }
    \label{pic:Simmoe}
\end{figure*}

\section{Cosine similarity of ICC}
We provide more experimental results on $ICC$. Figure~\ref{pic:Sim} represents the similarity of $ICC$ increments after full parameter fine-tuning, and Figure~\ref{pic:Simmoe} represents the ICC similarity of \textbf{CITI} method. The result demonstrates that \textbf{CITI} can significantly alleviates the Co-directional shift phenomenon.

\end{document}